\documentclass[times, review, 10pt]{elsarticle}

\usepackage{amssymb}
\usepackage{amsmath}
\usepackage{algorithm}
\usepackage{algorithmic}
\usepackage{booktabs}
\usepackage{multirow}
\usepackage{xcolor}
\usepackage[hidelinks]{hyperref}
\definecolor{darkgreen}{RGB}{0,100,0}

\journal{Pattern Recognition}

\begin{document}

\begin{frontmatter}



\title{Edit-Your-Interest: Efficient Video Editing via Feature Most-Similar Propagation} 


\author[label1,label2]{Yi~Zuo\corref{cor1}} 
\ead{yiiizuo@163.com}
\author[label1,label2]{Zitao~Wang} 
\author[label1,label2]{Lingling~Li\corref{cor1}} 
\ead{llli@xidian.edu.cn}
\author[label1,label2]{Xu~Liu} 
\author[label1,label2]{Fang~Liu} 
\author[label1,label2]{Licheng~Jiao}

\affiliation[label1]{organization={Key Laboratory of Intelligent Perception and Image Understanding of Ministry of Education},country={China}}
\affiliation[label2]{organization={School of Artificial Intelligence, Xidian University},
            addressline={No.2 Taibai South Road},
            city={Xi'an},
            postcode={710071},
            state={Shaanxi},
            country={China}}

\cortext[cor1]{Corresponding author}


\begin{abstract}
     Text-to-image diffusion models have recently demonstrated significant progress in video editing.
    However, existing video editing methods are severely limited by their high computational overhead and memory consumption. 
    Furthermore, these approaches often sacrifice visual fidelity, leading to undesirable temporal inconsistencies and artifacts such as blurring and pronounced mosaic-like patterns. 
    To address these dual challenges and strike a balance between computational efficiency and visual fidelity, we propose Edit-Your-Interest, a lightweight, text-driven, zero-shot video editing method.
    Edit-Your-Interest introduces a Spatio-Temporal Feature Memory bank (SFM) to cache spatial-attention features from previous frames under a fixed memory budget.
    The SFM maintains a representative candidate pool, reducing the computational overhead of full-sequence spatio-temporal modeling.
    To query this memory, we propose Feature Most-Similar Propagation (FMP), a similarity-based retrieval rule that propagates relevant cached tokens to the current frame.
    Together, SFM and FMP form a bounded-memory retrieval-based propagation design that improves temporal consistency while reducing blurring and mosaic-like artifacts.
    Finally, we introduce a feature-diversity-based SFM update algorithm to keep the memory representative and non-redundant throughout the video sequence.
    Furthermore, to enable precise object editing, we leverage cross-attention maps to automatically extract masks for the instances of interest.
    These masks are seamlessly integrated into the diffusion denoising process, enabling fine-grained control over target objects and allowing Edit-Your-Interest to perform highly accurate edits while robustly preserving the background integrity.
    Extensive experiments demonstrate that Edit-Your-Interest outperforms state-of-the-art methods in both efficiency and visual fidelity.
    The project page is available at \href{https://edit-your-interest.github.io/}{https://edit-your-interest.github.io}.
\end{abstract}




\end{frontmatter}



\section{Introduction}
In recent years, diffusion models have made significant progress in both text-to-image (T2I) and text-to-video (T2V) generation~\cite{hong2023cogvideo}, with notable models such as DALL·E, DiT~\cite{peebles2023scalable}, and Stable Diffusion~\cite{rombach2022high}.
In T2V generation, text-driven video editing models have attracted considerable attention due to their practical utility.

These models aim to generate edited videos that align with the description in the target prompt, conditioned on the source video, source prompt, and target prompt. 
The generated video should preserve the structural consistency of the source video.
Current text-driven video editing models generally fall into two main paradigms: fine-tuning-based~\cite{wu2023tune, zhong2025deco} and zero-shot video editing methods~\cite{qi2023fatezero}.
Fine-tuning-based methods adapt pre-trained models using video datasets or a single input video.

They incur training costs during model development or per-video adaptation.
In contrast, zero-shot video editing models offer a more resource-efficient alternative.
Therefore, we focus on text-driven zero-shot video editing to minimize resource usage while maintaining high-quality editing capabilities.

Existing zero-shot video editing models primarily rely on textual prompts to guide the editing of video content. 
Some approaches~\cite{hertz2023prompttoprompt, qi2023fatezero} integrate various attention maps during the inversion and sampling processes. 
However, these methods suffer from significant limitations when applied to long video sequences, as storing the global attention maps leads to excessive memory consumption and computational overhead.
Other approaches~\cite{geyer2023tokenflow, li2024video, kara2024rave} reduce attention map storage demands through keyframe sampling and sliding-window strategies. 
While promising, these methods often generate videos with low visual fidelity due to feature smoothing, which manifests as blurring and mosaic-like artifacts.
To achieve lightweight and practical video editing with high visual fidelity at low computational cost,
we propose two key ideas: (1) caching features from previous frames in a feature memory, and (2) propagating these cached features to the current frame.

In this paper, we propose Edit-Your-Interest, a lightweight zero-shot video editing framework that achieves high efficiency and visual quality simultaneously.
First, we introduce a Spatio-Temporal Feature Memory bank (SFM) to cache spatial-attention features from previous frames under a fixed memory budget.
The SFM retains a representative candidate pool, thereby avoiding the high computational overhead of global temporal attention.
Second, we propose Feature Most-Similar Propagation (FMP) as a similarity-based querying rule over this memory pool.
Together, SFM and FMP form a bounded-memory retrieval-based propagation design for efficient zero-shot video editing.
Third, we design an SFM update algorithm based on feature diversity to keep the memory representative and non-redundant across the video sequence.

Additionally, to enable precise object-level editing, we automatically extract masks for objects of interest from cross attention maps guided by the textual prompt, and seamlessly integrate them into the diffusion denoising process. 
This strategy supports fine-grained object editing without requiring external video segmentation models, while robustly preserving background integrity.

To summarize, our key contributions are as follows:
\begin{itemize}
	\item We propose Edit-Your-Interest, a lightweight zero-shot video editing framework that achieves high-quality editing with low computational overhead.
	\item We introduce a bounded-memory retrieval-based propagation design. SFM maintains a representative candidate pool under a fixed memory budget, and FMP queries this pool through similarity-based retrieval.
	\item We design a feature-diversity-based SFM update algorithm to keep the memory representative and non-redundant over long video sequences.
	\item For precise object editing, we automatically extract masks for objects of interest from cross attention maps and seamlessly integrate them into the diffusion denoising process.
	\item Our approach can process over 100 video frames on an RTX 4090 GPU with 24 GB of memory, demonstrating its practical efficiency and scalability. Moreover, our method achieves state-of-the-art editing performance on different videos, validating its effectiveness and generalization capability.
\end{itemize}

The remainder of this article is organized as follows. Section~\ref{sec:ii} covers related work. Section~\ref{sec:iii} details our proposed Edit-Your-Interest. Section~\ref{sec:iv} describes the experimental settings, comparative results, modules analysis, ablation studies and limitations. Section~\ref{sec:v} presents conclusions and future directions.

\begin{figure*}[t]
    \centering
    \includegraphics[width=\textwidth]{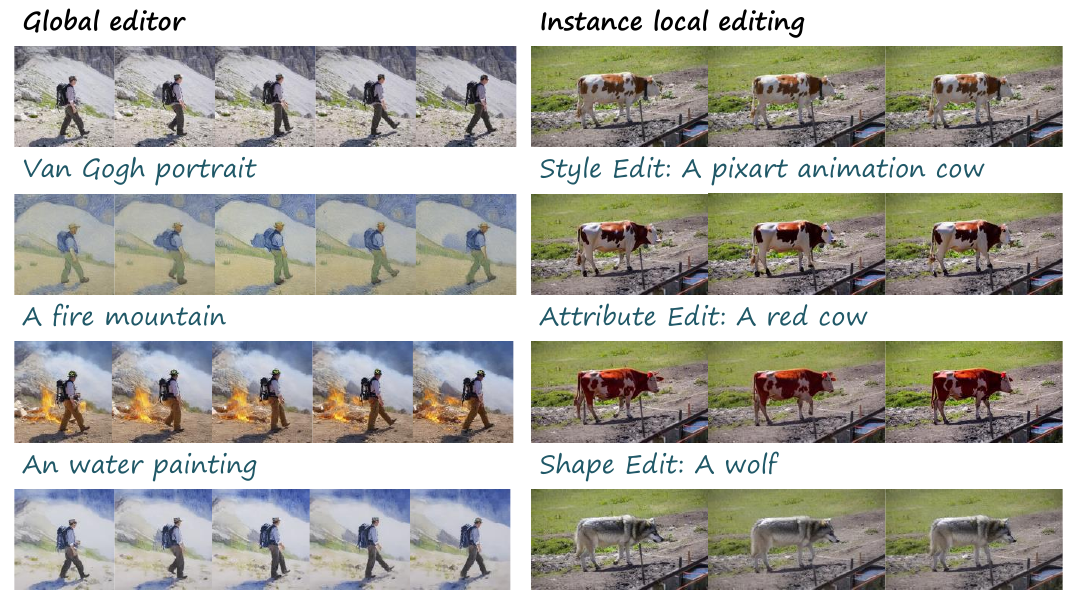}
    \caption{We propose Edit-Your-Interest, a zero-shot video editing method that supports both low-cost global editing (left) and precise instance local editing (right), while effectively preserving the entire background.}
    \label{fig:img1}
\end{figure*}

\section{RELATED WORK}\label{sec:ii}
The research areas most relevant to our method are text-driven image generation and editing, text-driven video editing with fine-tuning, and text-driven video editing with zero-shot.

\subsection{Text-driven image generation and editing}
Text-driven image generation diffusion models~\cite{peebles2023scalable, CHEN2026112583, CHANG2026111934} have become the dominant paradigm in image generation, owing to their remarkable ability to generate high-quality images. Among these approaches, Stable Diffusion~\cite{rombach2022high} stands out as the most prominent and has become a cornerstone pre-trained model in the field.

Image editing, an essential subfield of image generation, has similarly attracted significant research interest. In contrast to image generation, text-driven image editing models aim to modify the content of a given source image while preserving its original structure and layout.

P2P~\cite{hertz2023prompttoprompt} observed that cross attention layers play a critical role in controlling the relationship between the image's spatial layout and individual words in the prompt. It proposes a method to control image generation solely by editing the textual prompt.
PnP~\cite{ju2024pnp} corrects the inversion error by decoupling the source and target branches and minimizing the distance between them, thereby improving the fidelity of the edited image.
Instructpix2pix~\cite{brooks2023instructpix2pix} automates the construction of triplet-based image editing datasets, reframing editing tasks from cumbersome image descriptions into intuitive instruction following.
Eta~\cite{kang2025eta} designs an optimal $\eta$ function that is conditioned on time and region for diffusion inversion in the Denoising Diffusion Implicit Model (DDIM), with the goal of enhancing text-driven editing capability of real images.

A common strategy in image editing models~\cite{WANG2026112041, LIU2026112510} is to preserve the structural features of the source image by exchanging features between the source and target branches, while maintaining high editing fidelity under the guidance of the target prompt. However, a key limitation of these models is their inability to incorporate temporal information. As a result, although they achieve strong performance on image editing tasks, their application to video editing often results in noticeable temporal inconsistencies between adjacent frames.

\subsection{Text-driven video editing with fine-tuning}
The key difference between video editing and image editing lies in the input condition: instead of a single image, the input is a temporally coherent video sequence. Therefore, text-driven video editing methods, building upon image editing, must ensure that the generated edited video maintains inter-frame consistency.

Existing video editing methods can be broadly classified into two paradigms: fine-tuning-based and zero-shot video editing.
Fine-tuning can be performed on video datasets or a single input video, and dataset-level training need not be repeated for each new video.

Training-based video editing methods~\cite{ esser2023structure, singer2024video} improve temporal consistency by integrating spatio-temporal layers into the U-Net~\cite{ronneberger2015u} and fine-tuning them on large-scale video-text paired datasets.
However, because of the challenges in acquiring large-scale text-video paired datasets and the high computational overhead of training, these methods are often impractical for many application scenarios.

To mitigate this limitation, Tune-a-video~\cite{wu2023tune} proposes one-shot video editing, which loads the weights of a pre-trained T2I model and fine-tunes specific network layers on a single target video. 
$EI^2$~\cite{zhang2023towards} observed that directly adding temporal layers introduces covariate shift in the feature space. Therefore, it achieves effective editing via a feature distribution correction and interactive mechanism between fine and coarse information.
Stablevideo~\cite{chai2023stablevideo}, in contrast, introduces a Neural Layered Atlas (NLA) to decompose the video into foreground and background atlases, and then employs an aggregation network to preserve the geometric and appearance consistency of the edited object.
VMC~\cite{jeong2024vmc} fine-tunes only the temporal attention layer in a one-shot setting and introduces a motion distillation loss function to obtain the motion vectors that trace motion trajectories in the target video.
PropFly~\cite{Seo_2026_CVPR} learns edit propagation from supervision generated on the fly by a pre-trained video diffusion model.
UNIC~\cite{ICLR2026_2e37e565} unifies editing tasks through joint in-context modeling of source-video, noisy-latent, and conditioning tokens.
LIVEditor-14B~\cite{shao2026liveditorb} reduces attention cost through sparse context selection and query-adaptive computation.
LiveEdit~\cite{wang2026liveedit} targets real-time streaming editing through causal adaptation, few-step distillation, and mask caching.

One-shot editing incurs per-video fine-tuning costs; our method requires no additional training.

\pagebreak[4]\subsection{Text-driven video editing with zero-shot}

Zero-shot editing requires no additional model training or fine-tuning. We group these methods into three directions: attention/guidance and local control, feature/token reuse, and stronger pre-trained video priors.

The first direction uses attention manipulation, auxiliary guidance, or local control to improve editing consistency.
FateZero~\cite{qi2023fatezero} models inter-frame relationships by fusing attention maps across the denoising process. Ground-A-Video~\cite{jeong2024groundavideo} further incorporates geometric cues such as depth and optical flow to preserve structural coherence. DMT~\cite{yatim2024space} exploits motion priors from pre-trained T2V models and guides generation through space-time diffusion features, thereby better preserving scene layout and motion dynamics. More recent zero-shot works continue this line from new backbone perspectives. For example, DIVE~\cite{huang2025dive} explores subject-driven video editing with DINO-based semantic guidance, while QK-Edit~\cite{shen2025qk} revisits attention-based injection for MM-DiT backbones and shows that training-free attention manipulation remains effective for both image and video editing. 
VideoGrain~\cite{yang2025videograin} enables multi-grained video editing by modulating space-time attention.
PRIMEdit~\cite{teodoro2026primedit} uses instance-specific attention redistribution and disentangled sampling to localize multi-object edits and limit cross-instance leakage.
TraFrCo~\cite{wang2026training} decomposes videos using pre-trained segmentation and propagates background information across frames to reconstruct occluded regions.

A second line of work improves efficiency and consistency through temporal reuse of features or tokens. RAVE~\cite{kara2024rave} enhances coherence with randomized noise shuffling, and TokenFlow~\cite{geyer2023tokenflow} establishes correspondences across source frames and propagates edited keyframe features to non-keyframes via weighted interpolation. VidToMe~\cite{li2024vidtome} further shows that temporally redundant self-attention tokens can be aligned and merged across frames, yielding more temporally consistent and efficient zero-shot video editing. STEM~\cite{li2024video} reduces inversion cost by representing an entire video with a compact set of shared low-rank bases, thereby learning a unified spatio-temporal representation for all frames. 
TSPVE.~\cite{liu2026beyond} preserve temporal structure through semantic clip partitioning, anchor-guided token merging, and cross-clip feature fusion.
HetCache~\cite{Liu_2026_CVPR} reduces redundant attention computation by selectively caching context tokens across denoising timesteps.
Recent work such as GenProp~\cite{liu2025generative} also highlights the effectiveness of propagation-based designs for transferring edits through a video sequence.

A third direction leverages stronger pre-trained video generation priors while still preserving the zero-shot setting. FADE~\cite{zhu2025fade} introduces frequency-aware factorization for training-free video editing and shows that different components of video diffusion models can be modulated according to their temporal and spatial roles. 
StreamEdit~\cite{jiao2026streamedit} uses source-conditioned noise-to-target generation with few-step streaming models for efficient long-video editing.

Our method is most closely related to propagation-based zero-shot video editing methods. Instead of relying on weighted interpolation, token merging, or global low-rank inversion, we construct a Spatio-Temporal Feature Memory bank to cache spatial-attention features from previous frames and introduce Feature Most-Similar Propagation to retrieve and propagate the most similar feature tokens to the current frame. This design better preserves sharp local structures and alleviates the blurring or mosaic-like artifacts that may arise from feature averaging or token merging. 
At the same time, our method remains lightweight and training-free, making it suitable for efficient zero-shot video editing on long sequences.

\section{METHOD}\label{sec:iii}
To provide a detailed introduction to our proposed method, Section~\ref{sec:iii-a} reviews the preliminaries involved in video editing. 
Section~\ref{sec:iii-b} presents the overall architecture of our proposed Edit-Your-Interest. 
Section~\ref{sec:iii-c} introduces SFM and its update algorithm. 
Section~\ref{sec:iii-d} details our proposed FMP algorithm. 
Finally, Section~\ref{sec:iii-e} describes the automated extraction and injection strategy for masks of objects of interest.

\begin{figure*}
    \centering
    \includegraphics[width=0.78\textwidth]{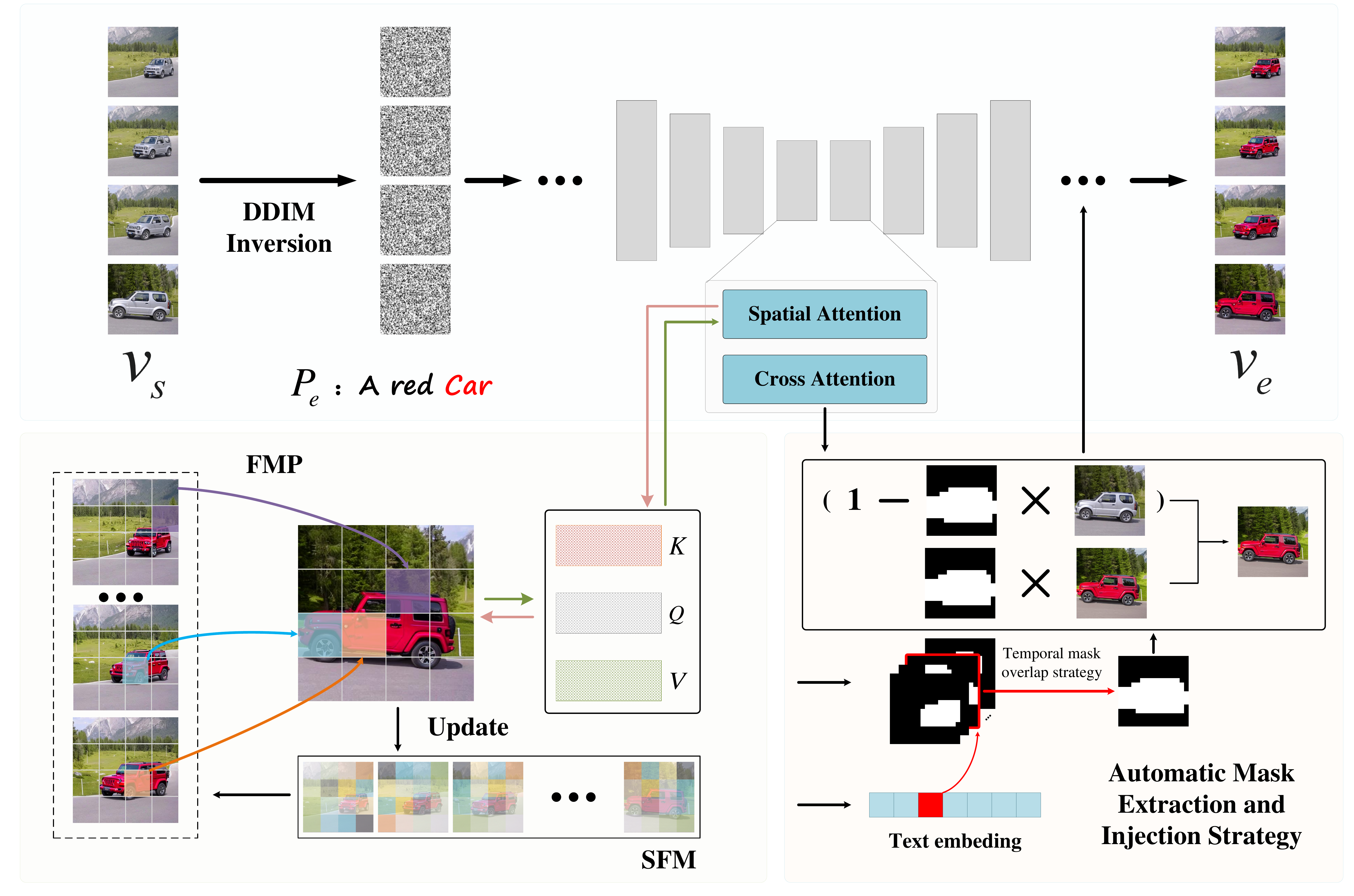}
    \caption{The pipeline of Edit-Your-Interest. (Top) We employ DDIM inversion to obtain the initial latent noise and then denoise the sequence via DDIM sampling. 
    (Bottom left) We construct a SFM to cache frame feature tokens. Subsequently, FMP propagates them to the current frame.
    (Bottom right) We introduce an Automatic Mask Extraction and Injection Strategy. This in-diffusion injection effectively suppresses boundary artifacts between foreground and background regions.
    }
    \label{fig:img3}
\end{figure*}

\subsection{Preliminaries}\label{sec:iii-a}
\textbf{DDPM and DDIM with Latent Diffusion Models.}
Denoising diffusion probabilistic models (DDPMs)~\cite{ho2020denoising} map the input noise $\boldsymbol{x}_T\sim\mathcal{N}(0,I)$ to clean samples $\boldsymbol{x}_{0}\sim q$ through an iterative denoising process. However, performing denoising directly in the pixel space requires significant computational overhead.
To improve the efficiency, latent diffusion models (LDMs)~\cite{rombach2022high} transfer the diffusion process from the pixel space to a lower-dimensional latent space through an autoencoder (VAE)~\cite{kingma2014auto}. Specifically, the encoder $\mathcal{E}$ of the VAE compresses an image $x$ into a low-resolution latent representation $z = \mathcal{E}(x)$, which is finally reconstructed back to image $\mathcal{D}(z) = x$ by the decoder $\mathcal{D}$.

During the forward diffusion process in the latent space, noise is iteratively added to the initial latent $z_0$ to obtain the noisy latent $z_t$ at timestep $t$:
\begin{equation}
    q(z_t|z_{t-1})=\mathcal{N}(z_t;\sqrt{1-\beta_t}z_{t-1},\beta_t\mathbf{I}),
\end{equation}
where $t \in \{1, \cdots, T\}$ is the current timestep, $z_t$ is the latent noise at timestep $t$. $\beta_t$ is sampled from a standard normal distribution. 

The backward process is the posterior probability distribution of the forward process, which can be obtained by derivation from Bayes' rule:
\begin{equation}
    \begin{aligned}
        p_\theta(z_{t-1}|z_t)=\mathcal{N}(z_{t-1};\mu_\theta(z_t,t),\Sigma_\theta(z_t,t)).
    \end{aligned}
\end{equation}

Since the clean image $x_0$ is unavailable during inference, we introduce the denoising network U-Net $\varepsilon_\theta$ to estimate the noise $\varepsilon$ added during the forward diffusion process. This is achieved by minimizing the following function:
\begin{equation}
    \min_\theta E_{x \sim q(x),\varepsilon\sim N(0,I),t}\left\|\varepsilon-\varepsilon_\theta\left(z_t,t,p\right)\right\|_2^2,
\end{equation}
where $p$ denotes the input prompt text and $z_t = \sqrt{\bar{\alpha}_t}z_0 + \sqrt{1-\bar{\alpha}_t}\varepsilon$ is the noisy latent at timestep $t$.

After training $\varepsilon_\theta$, deterministic DDIM inversion~\cite{song2021denoising} can be used to invert a real image into a noisy diffusion latent, while DDIM sampling accelerates the backward process. Both follow the same update rule:
\begin{equation}
    z_{t'}=\sqrt{\frac{\alpha_{t'}}{\alpha_{t}}}z_{t}+\left(\sqrt{\frac{1-\alpha_{t'}}{\alpha_{t'}}}-\sqrt{\frac{1-\alpha_{t}}{\alpha_{t}}}\right)\epsilon_{\theta}(z_t,t,p),
\end{equation}
where $\alpha_{t} = \prod^{t}_{s=1} (1-\beta_s)$ is the cumulative signal-to-noise ratio parameter in the noise schedule, and $t'=t-1$ for sampling or $t'=t+1$ for inversion.

\textbf{Video Editing with Diffusion Models.} Existing video editing models use DDIM inversion to map the source video $v_s$ to a noisy latent $z_T^{v}$ via its clean latent $z_0^{v} = \mathcal{E}(v_s)$:
\begin{equation}
    z_T^{v} = \mathrm{DDIM\_INV}(z_0^{v}, p_s),
\end{equation}
where $p_s$ is the prompt of the source video $v_s$, and $z_T^{v}$ is the inverted noisy latent of $z_0^{v}$. 

The edited video is then generated by iteratively denoising $z_T^{v}$ under the guidance of the editing prompt $p_e$:
\begin{equation}
    z_T^{v}\rightarrow \hat{z}_{T-1}^{v}\rightarrow \cdots \rightarrow \hat{z}_0^{v},
\end{equation}
where $\hat{z}_0^{v}$ is decoded to yield the edited video $v_e= \mathcal{D}(\hat{z}_0^{v})$. The inversion and denoising steps are designed to maximize the preservation of structural and semantic features from the source video $v_s$.

\subsection{Overall Architecture}\label{sec:iii-b}

Given an input video sequence $v_s =\{x_0, x_1, ... , x_n\}$, an input video text prompt $p_s$, and an editing text prompt $p_e$, where $x_i \in \mathbb{R}^{3\times H\times W}$ represents the $i$-th frame in $v_s$. Our goal is to edit the $v_s$ such that it aligns with $p_e$, generating the edited video $v_e$.
Specifically, we employ a pre-trained T2I diffusion model with robust generative capabilities as the backbone.
First, we initialize the noise sequence using DDIM inversion and then iteratively denoise it.

During the denoising process, we construct a Spatio-Temporal Feature Memory bank (SFM) to cache feature tokens from previous frames, significantly reducing the computational overhead of spatio-temporal modeling. we further introduce an SFM update algorithm that continuously refreshes feature tokens.

Subsequently, we propose a Feature Most-
Similar Propagation (FMP) method to propagate these cached features to the current frame, ensuring temporal consistency in the edited video and mitigating blurring and mosaic-like artifacts.

To enable precise editing of objects while preserving the background, we design an automated pipeline for extracting object masks of interest and seamlessly integrating them into the denoising process. This enables accurate editing of target objects without altering the background.
 
The overall pipeline of our method is illustrated in Fig.~\ref{fig:img3}.

\subsection{Spatio-Temporal Feature Memory bank}\label{sec:iii-c}

In video editing, a key objective is to ensure temporal coherence, often referred to as inter-frame consistency.
Existing zero-shot methods typically adopt one of two strategies: global feature propagation, which incurs high computational overhead, or keyframe feature propagation, where weighted averaging of features from adjacent keyframes often leads to blurring or mosaic-like artifacts (see Fig.~\ref{fig:img8}), and keyframe selection relies on manual intervention.

To reduce computational overhead while avoiding the artifact issues of keyframe interpolation, we introduce a Spatio-Temporal Feature Memory bank (SFM) that caches spatial feature tokens from the U-Net spatial attention layers. Formally:

\begin{equation}
    \mathcal{M} = \{sa_0, sa_1, \ldots, sa_L \},
\end{equation}
where $L$ denotes the length of SFM and $sa$ denotes the spatial feature tokens.

Storing all frames would be prohibitive. We therefore update the SFM dynamically to retain a diverse and representative subset. Specifically, we compute pairwise cosine distances among candidates, and iteratively remove the frame with the smallest distance to the rest, preserving feature diversity.

This update strategy can be viewed as a coverage-inspired redundancy-removal heuristic related to facility location. The goal is to keep a small memory that covers the temporal feature distribution while avoiding redundant frames. We use this view only as an intuition for the diversity-based update, rather than as a strict approximation guarantee. Compared with uniform sampling, which assumes constant temporal feature density, the diversity-based update adapts to the actual feature distribution.

\subsection{Feature Most-Similar Propagation}\label{sec:iii-d}

To maintain temporal consistency while reducing the artifacts caused by feature averaging, we propose Feature Most-Similar Propagation (FMP). FMP can be understood as the hard-attention limit of similarity-weighted propagation. In general, the propagated feature can be written as a softmax-weighted combination: $sa'_i = \sum_j w_j \cdot sa_j$ with $w_j = \exp(s_j / \tau) / \sum_k \exp(s_k / \tau)$, where $\tau > 0$ controls sharpness. As $\tau \to 0$, the weights degenerate to a one-hot vector $\mathbf{1}[j = \arg\max_k s_k]$, selecting only the most similar token. This formulation places FMP within the similarity-weighted propagation family, but it does not imply that hard selection is universally optimal. FMP operates at this limit, with a similarity threshold $\lambda$ as an additional safeguard: when no cached token exceeds $\lambda$, the original feature is retained unchanged.

Specifically, we first compute the cosine similarity between the current frame feature and each token in the SFM:
\begin{equation}
    \mathbf{s} = \mathbf{sa}_i^\top \left[ \mathbf{sa}_0, \mathbf{sa}_1, \ldots, \mathbf{sa}_{L-1} \right] ,
\end{equation}
where $\mathbf{sa}_i$ is the feature token of the current frame and $\mathbf{s}$ is the similarity vector. We then select the index of the most similar token,
\begin{equation}
    j^* = \arg\max_{j \in \{0, \cdots, L-1\}} s_j,
\end{equation}
and propagate it only if its similarity exceeds a threshold $\lambda$:
\begin{equation}
    \mathbf{sa}_i' =
    \begin{cases}
        \mathbf{sa}_{j^*}, & \text{if } s_{j^*} \geq \lambda, \\
        \mathbf{sa}_i, & \text{otherwise}.
    \end{cases}
\end{equation}

We present non-linear composability as an empirical motivation, not as a proven property of the U-Net feature space. It suggests that averaging mismatched tokens may cause blurring or mosaic-like artifacts, but it does not prove that hard top-1 selection is universally optimal. We therefore compare hard retrieval with softmax propagation under the same candidate pool in Section~\ref{sec:iiii-fpa}.

\begin{figure*}[t]
    \centering
    \includegraphics[width=0.91\textwidth]{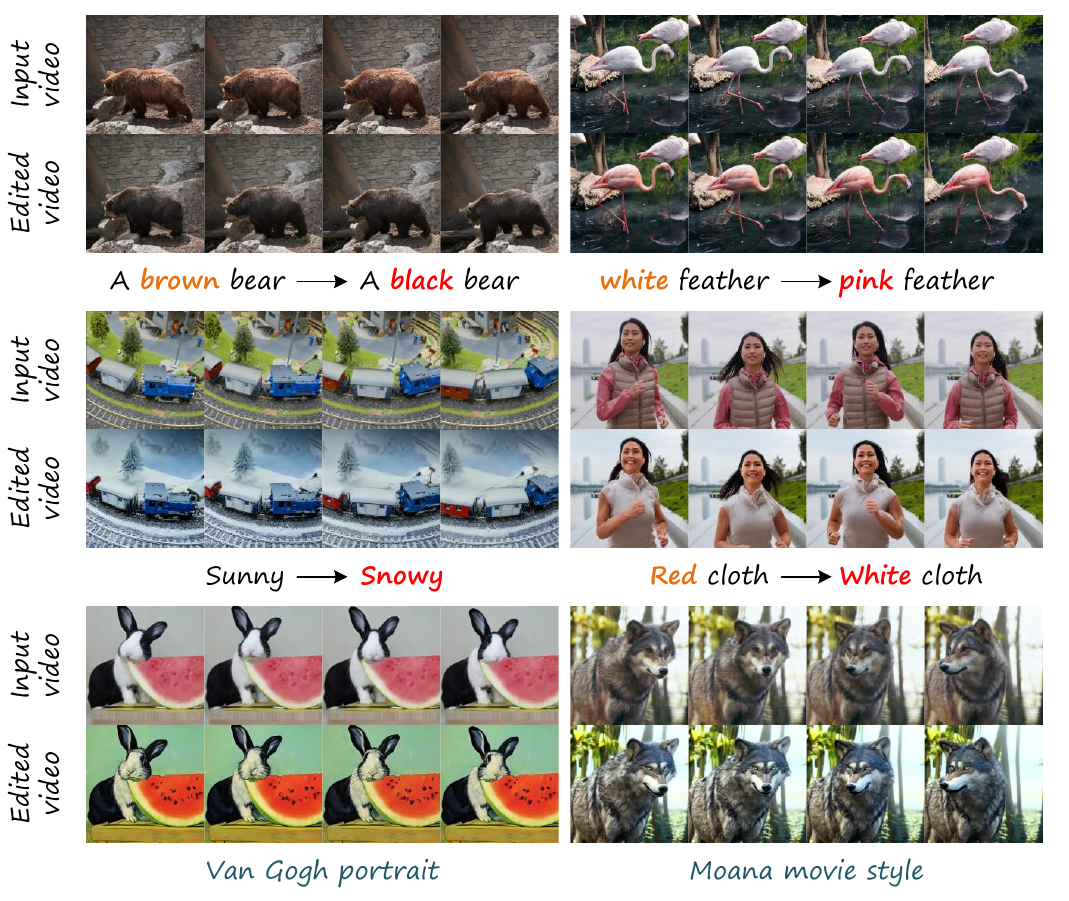}
    \caption{Examples of our method's results in instance object editing and global editing. As shown, our approach enables not only precise local object editing but also global editing.}
    \label{fig:img5}
\end{figure*}

\subsection{Automatic Mask Extraction and Injection Strategy}\label{sec:iii-e}

\begin{figure}[t]
    \centering
    \includegraphics[width=1\columnwidth]{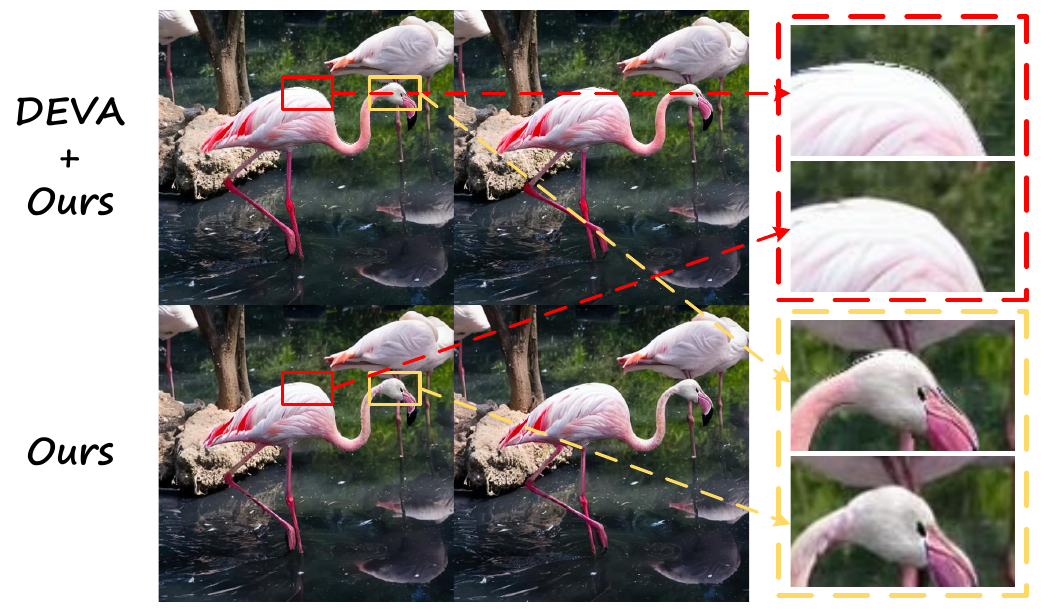}
    \caption{Comparison results between our method and direct replacement using semantic masks. It can be seen that our method mitigates boundary artifacts between foreground and background.}
    \label{fig:img17}
  \end{figure}

An intuitive idea for implementing instance-level object editing is to use semantic masks to replace corresponding regions in the input video. 
However, this strategy not only involves cumbersome steps (such as applying external video segmentation models), but also results in visible boundary artifacts between the foreground and background.

Inspired by~\cite{qi2023fatezero, hertz2023prompttoprompt}, we leverage cross attention maps to design a method for automatically extracting masks of objects of interest without relying on additional segmentation models. 
Moreover, our method achieves seamless blending between foreground and background regions, effectively eliminating visible seams (see Fig.~\ref{fig:img17}).

In the cross attention map, $K$, $V$, and $Q$ are Key, Value, and Query, respectively, where $K$ and $V$ are derived from textual features and $Q$ is obtained from spatial features. We compute the attention score map $\text{AttentionProb}=QK^T$, which represents the degree of association between the $Q$ and $K$. 
In simpler terms, $\text{AttentionProb}$ represents the similarity between textual words and spatial locations in the image. These similarity weights enable the alignment of semantic concepts in the text with visual elements in the image.

To extract instance masks from the cross attention map, we first identify the token index $w$ of the word in the prompt $p_s$ and construct a word selection vector $M_w = [\alpha_0, ..., \alpha_n]$, where $\alpha_i=1$ if $i=p$, and $\alpha_i=0$ otherwise. 
Next, we compute the instance mask $M_{ins}$ based on $\text{AttentionProb}$ and $M_w$:
\begin{equation}
    M_{ins} = (\text{AttentionProb} \times M_w) > \tau,
\end{equation}
where $M_{ins} \in \mathbb{R}^{H\times W}$ is a binary mask, and $\tau$ denotes a predefined threshold.
In our experiments, we observe that missing regions occasionally appear within the masks derived using single frames, which may be attributed to the strong feature coupling~\cite{yang2025videograin} of text features in the cross attention map (see Fig.~\ref{fig:img16}). 

To mitigate this, we propose a simple yet effective temporal mask overlap strategy. First, we extract the contour of the mask:
\begin{equation}
    c_{ins} = \text{contours}(M_{ins}),
\end{equation}
where $\text{contours}(\cdot)$ denotes the contour extraction operation. 
We then merge the contours from the current and previous frames and fill them to obtain a temporally consistent, robust instance mask:
\begin{equation}
    M^{\prime}_{ins} = \text{fill}(c^{i-1}_{ins} \cup c^{i}_{ins}).
\end{equation}
Finally, we inject the background features from the source video $v_s$ to preserve unedited regions:
\begin{equation}
    z^{\prime\prime}_t = M^{\prime}_{ins} \odot z^{\prime}_t + (1-M^{\prime}_{ins}) \odot z_t,
\end{equation}
where $z^{\prime}_t$ denotes the latent features of the edited video $v_e$, $z_t$ represents the latent features of the source video $v_s$, and $t \in [0.2T, T]$ indicates that background injection is applied during the later denoising steps to avoid interference with early semantic restructuring.
\begin{figure*}[t]
    \centering
    \includegraphics[width=0.88\textwidth]{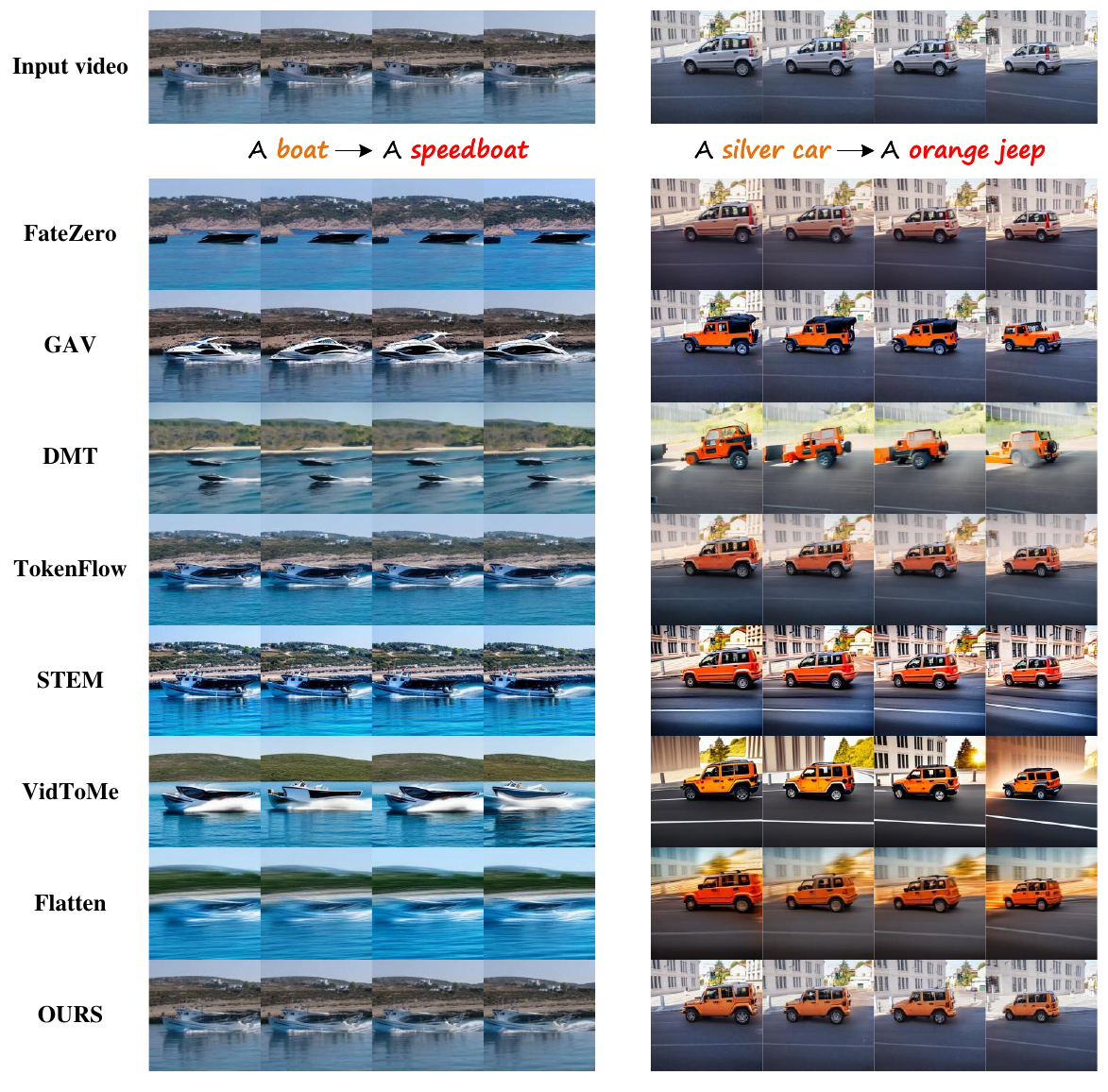}
    \caption{Qualitative Comparisons. The results of our proposed method and other state-of-the-art video editing methods are shown. It can be observed that our method not only successfully edits the target but also effectively preserves the background regions.}
    \label{fig:img6}
\end{figure*}

\section{EXPERIMENTAL}\label{sec:iv}

\subsection{Experimental Settings}
\subsubsection{Implementation Details} 
We adopt Stable Diffusion v1.5 with the official pre-trained weights as the baseline and use CLIP~\cite{radford2021learning} as the text encoder. To obtain the initial noise, we utilize DDIM inversion with $T=50$ steps, followed by DDIM sampling for the denoising process. 
The classifier-free guidance scale is set to 7.5, the similarity threshold $\lambda$ is set to 0.9, and the SFM length is set to 5 in all our experiments. On an RTX 4090 with 24GB of memory, our method can process videos containing up to 100 frames at a resolution of $512 \times 512$.

\subsubsection{Experimental Dataset} 
We evaluate our method on videos collected from the DAVIS~\cite{perazzi2016benchmark} and TGVE~\cite{wu2023cvpr} datasets. Each video contains 50 to 200 frames, cropped and resized to either $512 \times 512$ or $360 \times 640$ resolution.

\subsection{Comparison Methods and Evaluation Metrics}

\subsubsection{Comparison Methods} 
To demonstrate the superiority of our approach, we selected five state-of-the-art video editing methods for comparison: FateZero~\cite{qi2023fatezero}, Ground-A-Video (GAV)~\cite{jeong2024groundavideo}, TokenFlow~\cite{geyer2023tokenflow}, STEM~\cite{li2024video}, DMT~\cite{yatim2024space}, VidToMe~\cite{li2024vidtome} and Flatten~\cite{cong2024flatten}.

\paragraph{FateZero} 
FateZero proposes a framework for temporally consistent video editing that requires neither training on each target prompt nor user-provided masks. It achieves this by fusing and blending attention maps to preserve the original structure and motion information of the video.

\paragraph{Ground-A-Video (GAV)} 
GAV integrates spatially discrete textual grounding with spatially continuous geometric priors. It introduces a cross-frame gated attention, modulated cross-attention and optical flow guided inverted latents smoothing to achieve multi-attribute video editing.

\paragraph{TokenFlow} 
TokenFlow establishes feature correspondences across source video frames and propagates edited keyframe features to non-keyframes via weighted interpolation, thereby enforcing temporal consistency by preserving the source videos temporal structure.

\paragraph{STEM} 
STEM avoids per-frame DDIM inversion by representing the entire video with a shared set of low-rank bases (e.g., 256 bases). It optimizes these shared bases through an expectation-maximization iteration manner to obtain a unified spatio-temporal representation for all frames.

\paragraph{DMT} 
DMT converts the spatio-temporal features of the T2V diffusion model into spatial marginal mean (SMM) feature and guides new video generation through a new space-time feature, thereby achieving high-fidelity motion transfer across substantial structural differences.

\paragraph{VidToMe} 
VidToMe proposes a token merging strategy to align and merge temporally redundant self-attention tokens across frames, yielding more temporally consistent and efficient zero-shot video editing.

\paragraph{Flatten} 
Flatten introduces optical flow into the attention module to address the inconsistency issue. Specifically, it calculates optical flow and correlates patches along the same flow path across different frames to enhance visual consistency.

\begin{table*}[htbp]
  \centering
  \caption{Quantitative comparison of automatic metrics. \textbf{Bold} indicates the best result. Background preservation denotes background preservation, Foreground denotes foreground coherence, and Global denotes overall video quality.}
  \scalebox{0.7}{
    \begin{tabular}{c|c|cc|cc|cc|c}
    \toprule
    \toprule
    \multirow{2}[4]{*}{Method} & Text Alignment & \multicolumn{2}{c|}{Foreground} & \multicolumn{2}{c|}{Background preservation} & \multicolumn{2}{c|}{Global} & Fidelity \\
\cmidrule{2-9}          & CLIP-T & CLIP-F & Warp-err & SSIM  & PSNR  & CLIP-F & Warp-err & FID \\
    \midrule
    FateZero~\cite{qi2023fatezero} & 31.05 & 95.98 & 5.77  & 81.91 & 21.49 & 94.76 & 6.80  & 289.92 \\
    DMT~\cite{yatim2024space}   & 30.78 & \textbf{98.50} & 1.21  & 52.35 & 16.67 & 98.39 & \textbf{1.15} & 214.99 \\
    GAV~\cite{jeong2024groundavideo}   & 27.82 & 96.45 & 2.11  & 68.96 & 17.51 & 96.44 & 4.91  & 243.28 \\
    TokenFlow~\cite{geyer2023tokenflow} & 31.32 & 97.03 & 1.24  & 77.65 & 21.31 & 98.51 & 1.38  & 162.59 \\
    STEM~\cite{li2024video}  & 29.89 & 97.89 & 1.48  & 71.79 & 17.52 & 98.48 & 3.47  & 170.46 \\
    VidToMe~\cite{li2024vidtome} & 32.26 & 96.01 & 1.52  & 57.11 & 20.12 & 95.65 & 1.79  & 157.53 \\
    Flatten~\cite{cong2024flatten} & 30.23 & 97.12 & \textbf{1.05} & 62.54 & 15.16 & 97.49 & 1.57  & 208.32 \\
    \midrule
    Ours  & \textbf{32.29} & 97.92 & 1.19  & \textbf{86.52} & \textbf{29.26} & \textbf{98.94} & 1.21  & \textbf{120.12} \\
    \bottomrule
    \end{tabular}%
  }
  \label{tab:quantitative}%
\end{table*}%

\subsubsection{Evaluation Metrics}
To evaluate the effectiveness of our proposed Edit-Your-Interest, we assess the editing videos along four key dimensions: text alignment, temporal consistency, background preservation, and video fidelity, respectively. Additionally, we conduct a user study to measure perceptual quality.
\paragraph{Text alignment} 
We compute the CLIP similarity between the text prompt and each edited frame, averaged over the video, denoted as CLIP-T (scaled by $\times 100$).
\paragraph{Temporal consistency} 
We measure frame-to-frame coherence using two metrics: (1) the average CLIP similarity between adjacent frames (CLIP-F, scale by $\times 100$), and (2) the optical flow-based warping error following RAFT~\cite{teed2020raft} (Warp-Err, $\times 100$).
\paragraph{Background restoration} 
To quantify how well the background remains unchanged, we compute SSIM~\cite{wang2004image} and PSNR~\cite{hore2010image} between the background regions of the edited and source videos (both scaled by $\times 100$).
\paragraph{Video fidelity} 
We evaluate visual quality using the Fréchet Inception Distance (FID)~\cite{heusel2017gans} on generated frames.
\paragraph{User Study}
We invite 56 participants to rate the results on three criteria: Temporal Consistency (TC), Text Alignment (TA), and Visual Quality (Quality).

\subsection{Comparison Results}

\begin{table*}[htbp]
\centering
\caption{Comparison of Runtime and Memory Consumption. GPU denotes the GPU memory usage (in GB), RAM denotes the system memory usage (in GB), and Runtime indicates the time required to edite a video (in seconds). Values marked with an asterisk ($^{*}$) are adapted from~\cite{yang2025videograin}}
    \scalebox{0.75}{
    \begin{tabular}{c||ccc||ccc||ccc}
    \toprule
    \toprule
    \multirow{2}[4]{*}{Method} & \multicolumn{3}{c||}{8 frames} & \multicolumn{3}{c||}{16 frames} & \multicolumn{3}{c}{32 frames} \\
\cmidrule{2-10}          & GPU& RAM& Runtime & GPU & RAM & Runtime & GPU& RAM& Runtime\\
    \midrule
    FateZero~\cite{qi2023fatezero} & 18.57 & 71.13 & 154   & 27.34$^{*}$ & 144.21$^{*}$ & 517$^{*}$ & -     & -     & - \\
    GAV~\cite{jeong2024groundavideo} & 17.87 & 6.87 & 93    & 25.40 & 6.99 & 242   & 29.41 & 7.27 & 721 \\
    DMT~\cite{yatim2024space}   & 19.96 & 3.06 & 316   & 30.99 & 3.07 & 521   & 51.97 & 3.08 & 935 \\
    TokenFlow~\cite{geyer2023tokenflow} & 9.64 & 2.59 & 102   & 11.33 & 2.78 & 195   & 11.42 & 2.82& 403 \\
    STEM~\cite{li2024video}  & 9.78 & 2.84 & 52    & 11.46 & 2.90 & 97    & 11.57 & 2.93 & 198 \\
    OURS  & 9.38 & 2.81 & 71    & 9.86 & 2.90 & 126   & 10.96 & 2.92 & 252 \\
    \bottomrule
    \end{tabular}%
    }
\label{tab:addlabel3}%
\end{table*}%

\subsubsection{Qualitative Comparisons} 
We present the qualitative comparison results between Edit-Your-Interest and state-of-the-art methods in Fig.~\ref{fig:img6}.
Our method not only edits local instance object accurately according to textual prompts but also effectively preserves the background. This is attributed to our Automatic Mask Extraction and Injection Strategy, which enforces background consistency.
In contrast, TokenFlow and STEM achieve impressive editing results, but exhibit noticeable color shifts and saturation changes in the background. 
DMT and GAV fail to preserve the structural integrity of the source video. 
FateZero, while attempting to maintain the background through inversion-based masking, frequently produces edits that are misaligned with the textual prompt.
Overall, Edit-Your-Interest achieves precise, prompt-consistent instance-level video editing while maintaining high background fidelity.

\subsubsection{Quantitative Comparisons}
Quantitative comparison results are presented in Table.~\ref{tab:quantitative}. 
Our proposed Edit-Your-Interest achieves the best performance across all four metrics: text alignment (CLIP-T), temporal consistency (CLIP-F and Warp-err), background preservation (SSIM and PSNR), and video fidelity (FID). 
These results demonstrate that Edit-Your-Interest offers superior text controllability, temporal coherence, and higher editing quality compared to existing methods.

\begin{figure}[htbp]
	\centering
	\includegraphics[width=1\textwidth]{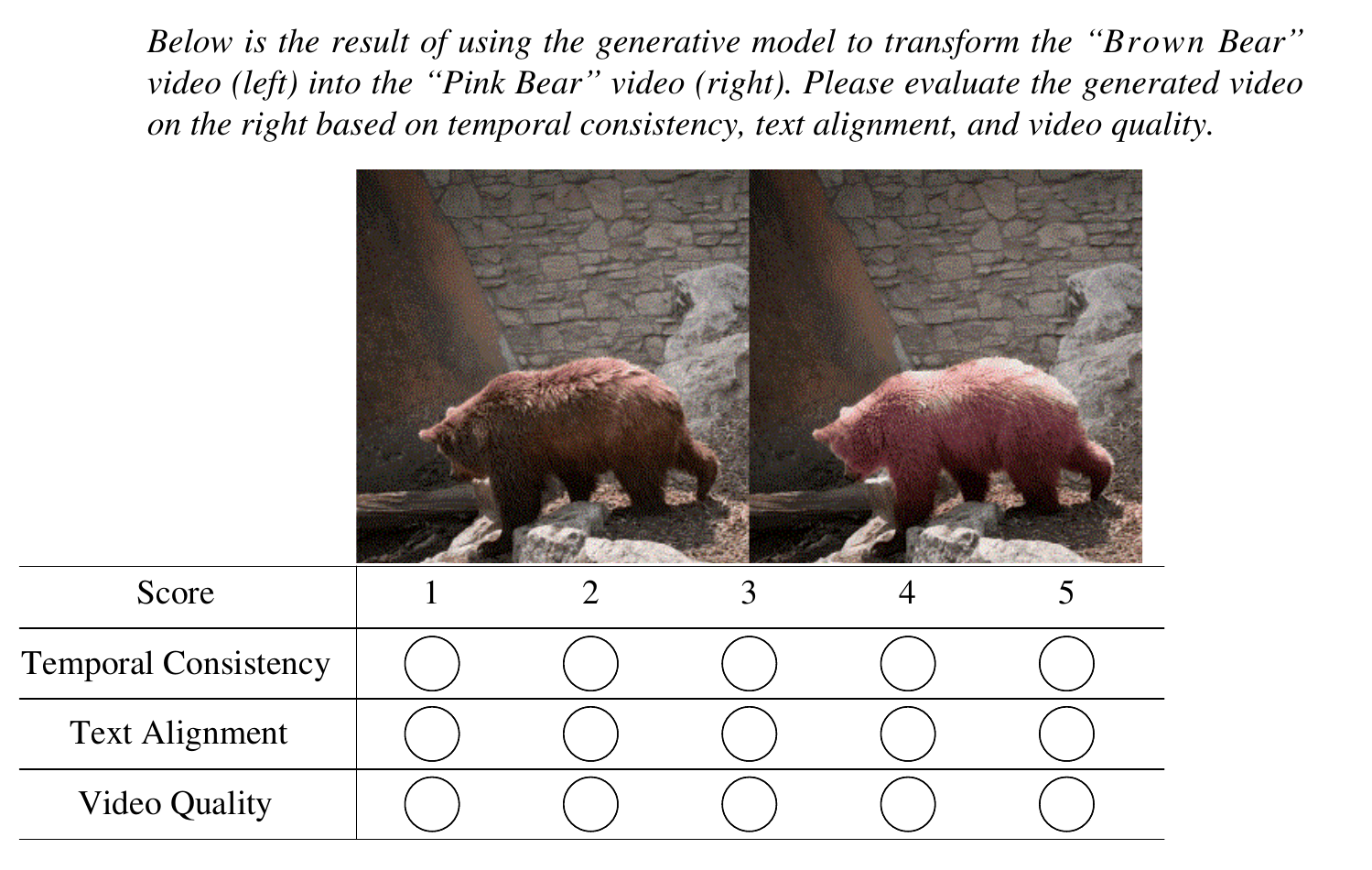}
	\caption{The example of the scoring interface for the user study. Participants are required to select a specific score from the provided table for each video.}
	\label{fig:img23}
  \end{figure}
\subsubsection{User Study}

To assess whether our method aligns with human perception, we conducted a user study with 56 participants from diverse backgrounds. 

Each participant was shown with the source video, the transform text prompt, and the edited videos from all methods. 
The presentation order of the videos was randomized, and method names were concealed to ensure unbiased evaluations.

Participants were asked to evaluate the generated videos based on three criteria: temporal consistency (TC), text alignment (TA), and video quality (Quality). 
The evaluation was conducted on a 5-point Likert scale, where 1 represents the lowest score and 5 represents the highest. 
After collecting all responses, we computed the average score per method by aggregating ratings across participants and videos. 
Finally, the user study score $S$ was derived by normalizing the aggregating scores with respect to the maximum possible score. The calculation of $S$ is as follows:
\begin{equation}
	S =\frac{\sum_{j \in J} \sum_{i=1}^{N} s_i^j}{5 \times length(J)},
\end{equation}
Where $s_i^j$ denotes the rating score assigned by the $i$-th participant to the $j$-th video, and $N$ represents the total number of participants. 

Fig.~\ref{fig:img23} shows a visualization of the interface that the participants can see. Table.~\ref{tab:userstudy} presents the results of the user study, demonstrating that our method best aligns with human perception across TC, TA, and quality metrics.

\begin{table}[tbp]
    \centering
    \caption{Representation-level analysis with temperature sweep. For all metrics except LV mean and GE, values closer to source indicate less artifact-induced distortion.}
    \begin{tabular}{c|cccccc}
    \toprule
    Method & LV (mean) & LV (std) & HFR & PSD & GE & DCT-E \\
    \midrule
    Source video & 413.50 & 32.30  & $4.54\times10^{-3}$ & 6.45 & 3.74 & 0.88 \\
    TokenFlow    & 596.53 & 132.99 & $7.23\times10^{-3}$ & 9.73 & 3.56 & 0.89 \\
    \midrule
    $\tau=1.0$   & 377.05 & 57.95  & $5.24\times10^{-3}$ & 6.75 & 3.61 & 0.84 \\
    $\tau=0.5$   & 367.94 & 57.08  & $5.15\times10^{-3}$ & 6.74 & 3.58 & 0.82 \\
    $\tau=0.1$   & 378.89 & 57.28  & $5.33\times10^{-3}$ & 6.75 & 3.60 & 0.85 \\
    $\tau=0.01$  & 379.58 & 58.06  & $5.45\times10^{-3}$ & 6.72 & 3.61 & 0.86 \\
    $\tau=0$ (Ours) & 350.92 & 56.11 & $5.34\times10^{-3}$ & 6.71 & 3.58 & 0.84 \\
    $\infty$ (uniform)      & 376.62 & 58.10  & $5.23\times10^{-3}$ & 6.75 & 3.61 & 0.84 \\
    \bottomrule
    \end{tabular}
    \label{tab:temp_sweep}
\end{table}

\subsection{Visual Results and Module Analysis}

\begin{figure}[t]
    \centering
    \includegraphics[width=0.8\columnwidth]{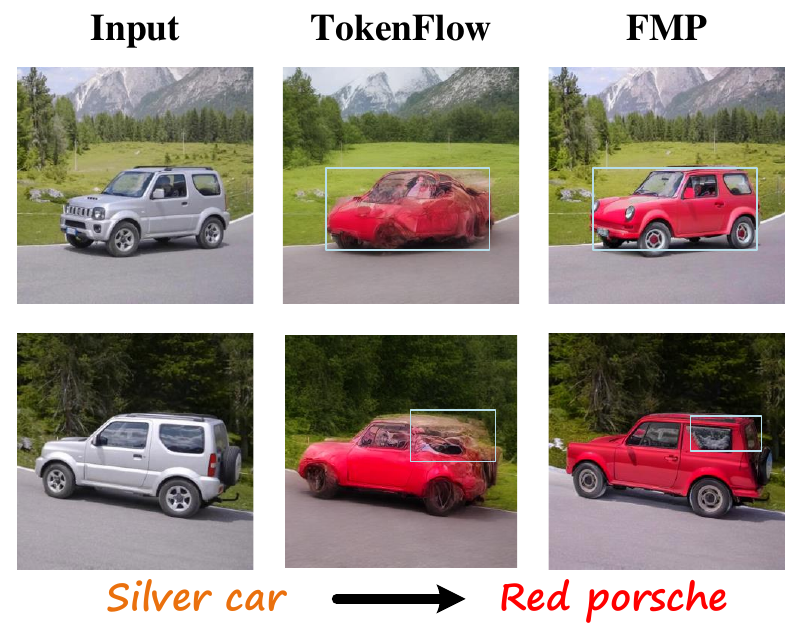}
    \caption{Comparison between FMP and interpolation-like weighted propagation methods. It can be observed that FMP mitigates blurring and mosaic-like artifacts, thereby enhancing the fidelity of edited videos.}
    \label{fig:img8}
  \end{figure}
\subsubsection{Visual Analysis} 
As shown in Fig.~\ref{fig:img1} and Fig.~\ref{fig:img5}, our Edit-Your-Interest supports instance-level editing of styles, attributes, and shapes. For example:

\paragraph{Style editing}
In the second row, the cow is transformed into a pixel-art animated cow.

\paragraph{Attribute editing}
In the third row, the cow's color is changed to red.

\paragraph{Shape editing}
In the fourth row, the cow is replaced with a wolf.

Notably, the background remains largely intact across all these edits, demonstrating Edit-Your-Interest's strong background preservation capability.
Moreover, Edit-Your-Interest also supports global editing. For example, in the left column of Fig.~\ref{fig:img1}, the video is transformed into the style of a Van Gogh portrait and an water painting, respectively.

\subsubsection{Feature Propagation Analysis}
\label{sec:iiii-fpa}
Fig.~\ref{fig:img8} compares FMP with TokenFlow, which is conceptually related to our propagation design. Our method produces sharper results and avoids blurring or mosaic-like artifacts because it queries a bounded SFM candidate pool by similarity-based retrieval, rather than averaging keyframe features. This separates candidate availability from the query rule.

To separate candidate-pool construction from propagation-rule choice, we compare keyframe and SFM candidates under different propagation rules in Table~\ref{tab:method_ablation}. Keyframe candidates are the two neighboring pivot/keyframe candidates used by TokenFlow, while SFM candidates additionally include retained memory features.

\begin{table}[t]
  \centering
  \small
  \caption{Quantitative comparison of candidate-pool construction and propagation strategies. Override denotes the fraction of tokens propagated from the candidate pool after passing the confidence threshold.}
  \label{tab:method_ablation}
  \setlength{\tabcolsep}{3.5pt}
  \resizebox{\linewidth}{!}{
  \begin{tabular}{lccccc}
    \toprule
    \multirow{2}{*}{Method}
    & \multicolumn{1}{c}{Text}
    & \multicolumn{2}{c}{Foreground}
    & \multicolumn{1}{c}{Fidelity}
    & \multirow{2}{*}{Override} \\
    \cmidrule(lr){2-2}
    \cmidrule(lr){3-4}
    \cmidrule(lr){5-5}
    & CLIP-T$\uparrow$
    & CLIP-F$\uparrow$
    & Warp-err$\downarrow$
    & FID$\downarrow$
    & \\
    \midrule
    Keyframe candidates + weighted
    & 31.32
    & 97.03
    & 1.24
    & 162.59
    & N/A \\
    Keyframe candidates + hard
    & 31.46
    & 97.15
    & 1.22
    & 144.75
    & 0.6832 \\
    SFM candidates + softmax ($\tau=0.5$)
    & 32.20
    & \textbf{98.14}
    & \textbf{1.18}
    & 124.03
    & 0.8050 \\
    SFM candidates + hard
    & \textbf{32.29}
    & 97.92
    & 1.19
    & \textbf{120.12}
    & 0.8031 \\
    \bottomrule
  \end{tabular}
  }
\end{table}

We further conduct a representation-level temperature sweep over softmax propagation (Table~\ref{tab:temp_sweep}) using Laplacian variance (LV), high-frequency ratio (HFR), mean power spectral density (PSD), gradient entropy (GE), and DCT entropy (DCT-E).

Table~\ref{tab:method_ablation} shows two effects. Hard retrieval improves over weighted interpolation under the same keyframe candidate pool, showing that the propagation rule matters. Replacing keyframe candidates with SFM candidates brings further gains, showing the benefit of the representative memory pool. Under SFM candidates, hard retrieval favors CLIP-T and FID, while softmax propagation slightly improves foreground temporal consistency.

\begin{figure}[t]
    \centering
    \includegraphics[width=0.8\columnwidth]{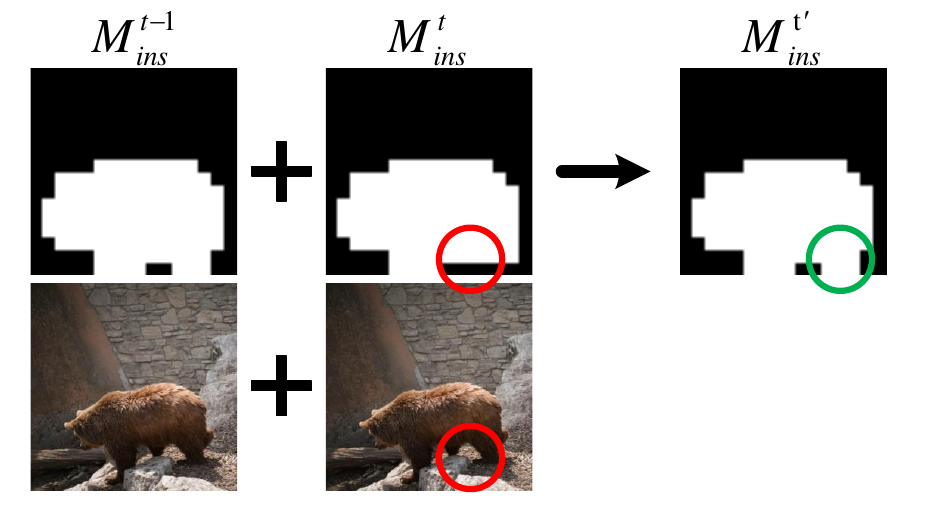}
    \caption{Visualization of temporal mask overlap strategy. This demonstrates that the temporal mask overlap strategy can effectively generate robust instance masks.}
    \label{fig:img16}
\end{figure}

\subsubsection{Temporal Mask Overlap Analysis}
In this section, we analyze the importance of our proposed temporal mask overlap strategy. In Edit-Your-Interest, instance masks are extracted from the cross attention layer maps. However, these masks are often incomplete, especially at the edges, which is likely due to the strong coupling in the text-to-image alignment process. 

To address this challenge, we generate robust instance masks by merging and filling the contours of masks from two consecutive frames.
As shown in Fig.~\ref{fig:img16}, the bear's foot initially exhibits a missing mask region in frame $t$. After applying our temporal mask overlap strategy, the occluded or fragmented part is effectively recovered, yielding a complete and robust coherent instance mask.

\begin{table}[tbp]
	\centering
	\caption{Hyperparameters Analysis: the similarity threshold $\lambda$, the mask threshold $\tau$, and the length of SFM (SFM-L).}
      \scalebox{0.8}{
	  \begin{tabular}{c|c||c|c||c|c}
	  \toprule
	  \toprule
	  $\lambda$ & Warp-err & $\tau$   & PSNR$^b$  & SFM-L & Warp-err \\
	  \midrule
	  0.7   & 3.92  & 0.2   & 28.17 & 1     & 1.57 \\
	  0.8   & 2.42  & 0.3   & 29.26 & 3     & 1.33 \\
	  0.9   & 1.19  & 0.4   & 29.03 & 5     & 1.19 \\
	  0.95  & 1.44  & 0.5   & 27.04 & 7     & 1.21 \\
	  \bottomrule
	  \end{tabular}%
      }
	\label{tab:addlabel4}%
  \end{table}%

\subsubsection{Hyperparameter Analysis.}
To determine the optimal values of the similarity threshold $\lambda$, the mask threshold $\tau$, and the length of SFM, we conducted a hyperparameter sensitivity analysis. 
Table.~\ref{tab:addlabel4} summarizes the metric results for different configurations. 

We observe that temporal consistency is maximized when $\lambda = 0.9$. The values that are too high restrict feature propagation by being overly selective, while values that are too low introduce noisy or irrelevant matches, degrading editing accuracy.
The best background preservation is achieved at $\tau = 0.3$, which is likely attributed to the limitations of the diffusion model's cross attention maps. 
Additionally, while temporal consistency is optimal when the length of SFM is set to either 5 or 7, we select 5 in our experiments to reduce computational storage requirements.

\begin{table*}[tbp]
  \centering
  \caption{The results of the user study. \textbf{Bold} indicates the best result.}
  \scalebox{0.8}{
    \begin{tabular}{c|ccc}
    \toprule
    \toprule
    Method & TC    & TA    & Quality \\
    \midrule
    FateZero & 67.50 & 73.5  & 71.75 \\
    DMT   & 72.75 & 69.75 & 79.75 \\
    GAV   & 71.75 & 81.50 & 69.50 \\
    TokenFlow & 88.75 & 88.00 & 80.95 \\
    STEM  & 87.25 & 84.00 & 84.00 \\
    OURS  & \textbf{90.25} & \textbf{95.5} & \textbf{90.75} \\
    \bottomrule
    \end{tabular}%
  }
  \label{tab:userstudy}%
\end{table*}%

\subsubsection{Efficiency Analysis.}
Runtime and memory consumption are critical metrics for evaluating video editing methods and represent key bottlenecks to their practical deployment. 
For a fair comparison, we evaluate these methods on an NVIDIA A800 GPU with 80 GB memory, a 14 vCPU Intel(R) Xeon(R) Gold 6348 CPU @ 2.60 GHz, and 100 GB RAM, using the default configurations from their official codes without modifications. 
The comparison results are reported in Table.~\ref{tab:addlabel3}. 
However, we were unable to conduct experiments beyond 16 frames for FateZero due to its excessive memory requirements.

While STEM achieves the lowest computational cost, it shows weaker performance in temporal consistency, text alignment, and background preservation, as evidenced by the quantitative results in Table~\ref{tab:quantitative}.
In contrast, our Edit-Your-Interest achieves state-of-the-art performance across all evaluation metrics while maintaining a low computational cost.

\begin{figure*}[t]
    \centering
    \includegraphics[width=1\textwidth]{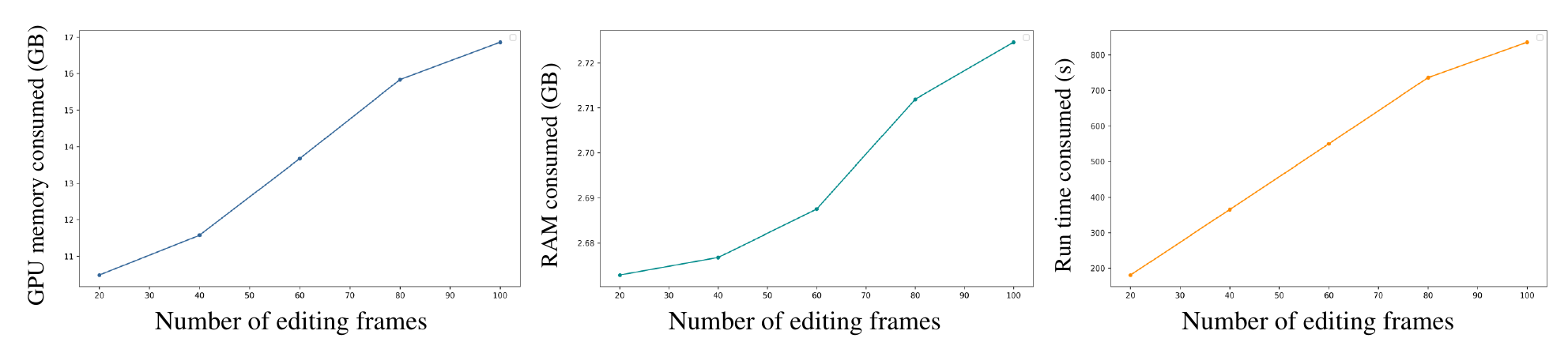}
    \caption{The computational overhead of our proposed Edit-Your-Interest when processing videos with varying numbers of frames.}
    \label{fig:img9}
\end{figure*}

Moreover, as shown in Fig.~\ref{fig:img9}, Edit-Your-Interest can efficiently edit videos with over 100 frames on a consumer-grade NVIDIA RTX 4090 GPU with 24 GB without consuming excessive RAM, demonstrating its scalability and practicality for real-world applications.

\begin{table}[htbp]
	\centering
	\caption{Ablation Study Results. PSNR$^b$ denotes PSNR computed on the background region.}
    \scalebox{0.8}{
	  \begin{tabular}{c|cc}
	  \toprule
	  \toprule
	  Method & Warp-err $\downarrow$ & PSNR$^b$ $\uparrow$ \\
	  \midrule
	  Baseline & 10.87     & 21.79 \\
	  w/o AMEIS & 3.12     & 21.89 \\
	  w/o FMP & 8.53     & 26.71 \\
	  OURS  & 1.19     & 29.26 \\
	  \bottomrule
	  \end{tabular}%
    }
	\label{tab:addlabel}%
  \end{table}%
\subsection{Ablation Study}
To validate the contributions of the Automatic Mask Extraction and Injection Strategy (AMEIS) and Feature Most-Similar Propagation (FMP) method to our overall framework, we adopt PnP-Inversion~\cite{ju2024pnp} as the baseline and perform ablation studies by individually disabling each component. The quantitative results are summarized in Table~\ref{tab:addlabel}. In addition to this component-level ablation, Section~\ref{sec:iiii-fpa} further separates the effects of candidate-pool construction and propagation-rule choice.

We observe that Automatic Mask Extraction and Injection Strategy plays a critical role in enabling precise instance-level editing while maximizing background fidelity by injecting unedited background features from the source input. In contrast, FMP significantly improves temporal consistency, leading to smoother and more consistent video generation across frames.

\subsection{Limitations and Future Work}

\begin{figure*}[t]
    \centering
    \includegraphics[width=0.8\textwidth]{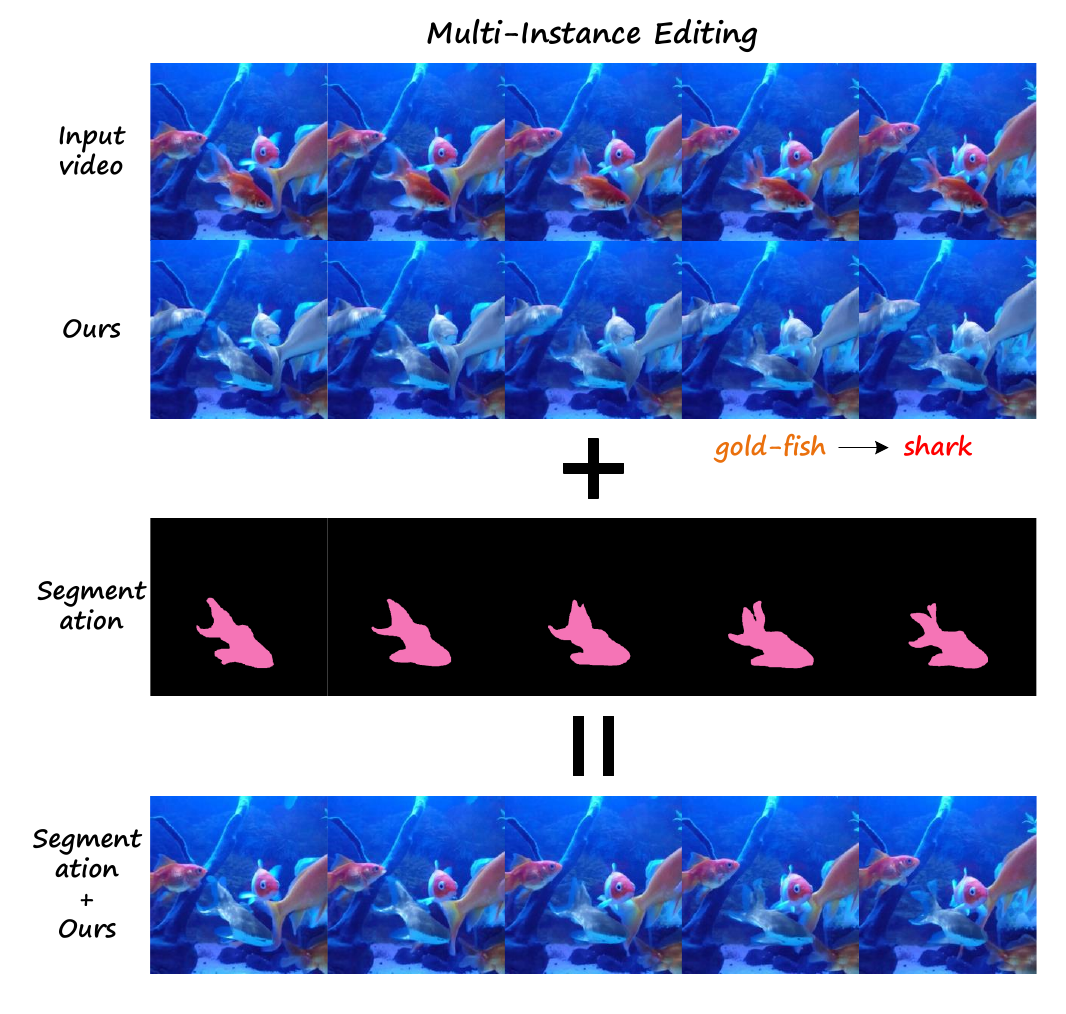}
    \caption{Our proposed Edit-Your-Interest achieves precise editing in multi-instance scenarios by integrating video instance segmentation model. As demonstrated, this method accurately edites specified target objects while preserving the background in complex scenes containing multiple instances of the same class.}
    \label{fig:img24}
\end{figure*}
\subsubsection{Multi-Instance Editing.} 
Since the instance masks in our method are extracted from the cross attention maps of the diffusion model, they lack the ability to disambiguate multiple instances of the same object class. 
To address this limitation, we propose integrating Edit-Your-Interest with a pre-trained video instance segmentation model. 

Specifically, the segmentation model provides instance-specific masks for objects of the same category, which are then incorporated into our method to enable targeted editing of individual instances.
Crucially, unlike methods that directly apply mask overlays to generate results, we inject the masks progressively during the diffusion denoising process. 
This in-diffusion integration effectively suppresses visible segmentation artifacts, particularly along object boundaries.

As illustrated in Fig.~\ref{fig:img24}, Edit-Your-Interest alone cannot distinguish the individual goldfish in a multi-instance scene. By leveraging an external segmentation model to extract the mask of the target goldfish and fusing it into the Edit-Your-Interest pipeline, we achieve precise, instance-level editing in complex multi-object videos. This suggests that extending AMEIS to other modalities, such as depth, segmentation, or optical flow guidance, is a promising future direction.

\section{Conclusion}\label{sec:v}
In this paper, we propose Edit-Your-Interest, a lightweight framework for zero-shot video editing, designed to address two major challenges in video editing: high computational overhead and visual blurring.
To mitigate computational overhead, we construct a Spatio-Temporal Feature Memory bank (SFM) that caches feature tokens from previous frames. We further design an update algorithm based on feature diversity to continuously refresh the SFM.
To mitigate visual blurring and mosaic-like artifacts, we propose the Feature Most-Similar Propagation (FMP) method, which propagates the most similar features from the SFM to the current frame via cross-frame similarity matching.
In addition, for instance-level object editing, we design an automated pipeline that extracts masks of objects of interest and seamlessly integrates them into the denoising process.
Overall, the framework combines representative memory construction, similarity-based feature retrieval, and mask-guided background preservation into a lightweight zero-shot video editing pipeline.
Extensive experiments demonstrate the effectiveness of our proposed Edit-Your-Interest.

\section{Acknowledgments}

This work was supported in part by the Joint Funds of the National Natural Science Foundation of China (U22B2054), the National Natural Science Foundation of China (62076192, 62276199, 62431020 and 62276201), the 111 Project, the Program for Cheung Kong Scholars and Innovative Research Team in University (IRT 15R53), the Science and Technology Innovation Project from the Chinese Ministry of Education.




 \bibliographystyle{elsarticle-num-names} 
 \bibliography{references}






\end{document}